\documentclass{article}
\usepackage{ijcai24}

\usepackage{times}
\usepackage{soul}
\usepackage{url}
\usepackage[hidelinks]{hyperref}
\usepackage[utf8]{inputenc}
\usepackage[small]{caption}
\usepackage{graphicx}
\usepackage{amsmath}
\usepackage{amsthm}
\usepackage{booktabs}
\usepackage{algorithm}
\usepackage{algorithmic}
\usepackage[switch]{lineno}
\usepackage{multirow}
\usepackage{amssymb}
\usepackage{bbding}
\usepackage{fontawesome}
\usepackage{xspace}
\usepackage{placeins}

\usepackage{colortbl}
\definecolor{LightCyan}{rgb}{0.88,1,1}
\definecolor{Gray}{gray}{0.92}

\newcommand{\eg}{\textit{e.g.}}
\newcommand{\ie}{\textit{i.e.}}
\newcommand{\our}{ENGINE\xspace}
\newcommand{\ours}{ENGINE\xspace}

\newcommand{\oure}{ENGINE~(Early)\xspace}
\newcommand{\oures}{ENGINE~(Early)\xspace}
\newcommand{\logo}{\includegraphics[scale=0.033]{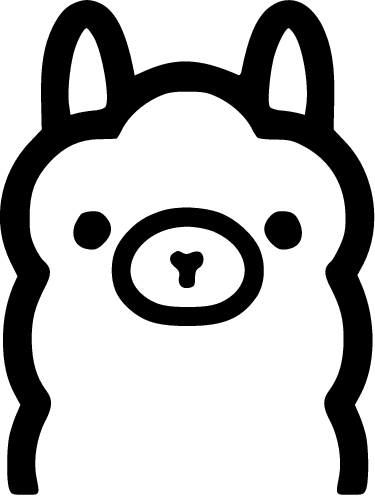}\xspace}

\DeclareMathOperator*{\argmin}{arg\,min}
\DeclareMathOperator*{\argmax}{arg\,max}

\title{Efficient Tuning and Inference for Large Language Models on Textual Graphs}

\author{
Yun Zhu$^{1,}$\footnotemark[1]
\and
Yaoke Wang$^{1,}$\footnotemark[1]\and
Haizhou Shi$^2$\And
Siliang Tang$^{1,}$\footnotemark[2]
\affiliations
$^1$Zhejiang University\\
$^2$Rutgers University
\emails
\{zhuyun\_dcd, wangyaoke\}@zju.edu.cn,
haizhou.shi@rutgers.edu,
siliang@zju.edu.cn
}

\begin{document}

\maketitle

\renewcommand{\thefootnote}{\fnsymbol{footnote}}
\footnotetext[1]{Equal Contribution}
\footnotetext[2]{Corresponding Author}

\begin{abstract}
Rich textual and topological information of textual graphs need to be modeled in real-world applications such as webpages, e-commerce, and academic articles. Practitioners have been long following the path of adopting a shallow text encoder and a subsequent graph neural network~(GNN) to solve this problem. In light of recent advancements in large language models (LLMs), it is apparent that integrating LLMs for enhanced textual encoding can substantially improve the performance of textual graphs. Nevertheless, the efficiency of these methods poses a significant challenge. In this paper, we propose ENGINE, a \emph{parameter- and memory-efficient fine-tuning method} for textual graphs with an LLM encoder. The key insight is to combine the LLMs and GNNs through a \emph{tunable side structure}, which significantly reduces the training complexity without impairing the joint model's capacity. Extensive experiments on textual graphs demonstrate our method's effectiveness by achieving the best model performance, meanwhile having the lowest training cost compared to previous methods. Moreover, we introduce two variants with caching and dynamic early exit to further enhance training and inference speed. Specifically, caching accelerates ENGINE's training by 12x, and dynamic early exit achieves up to 5x faster inference with a negligible performance drop~(at maximum 1.17\% relevant drop across 7 datasets). Our codes are available at: \href{https://github.com/ZhuYun97/ENGINE}{\texttt{https://github.com/ZhuYun97/ENGINE}}

\end{abstract}
\section{Introduction}
Textual graphs are pervasive in the real world, like academic networks~\cite{wang2020microsoft}, webpages~\cite{mernyei2020wiki} and e-commerce datasets~\cite{chiang2019cluster}. In the early stages, shallow embedding methods~\cite{mikolov2013distributed,zhang2010understanding} and graph neural networks~\cite{gcn,sage,gat} are combined to analyze and process this data like Figure~\ref{fig:comparsion}a. 
However, static shallow embedding methods struggle to capture context-aware information and complex semantic relationships, limiting their ability to exploit the richness of text attributes, particularly in graph tasks.
\begin{figure}
    \centering
    \includegraphics[scale=0.7]{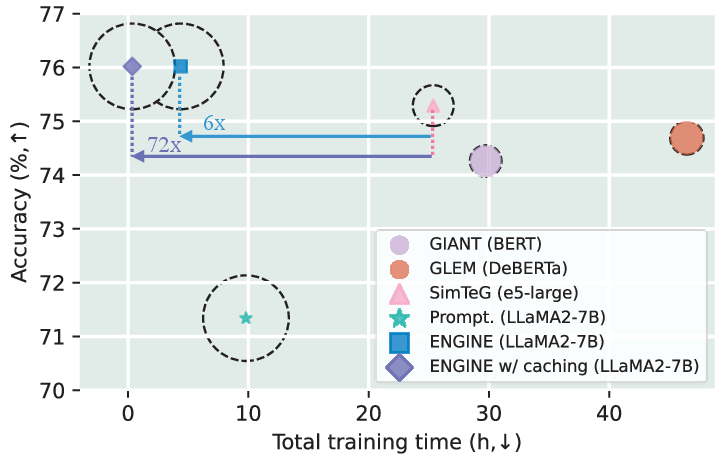}
    \caption{Comparison of performance and training efficiency between \ours and baselines on the large-scale textual graph dataset OGBN-ArXiv, where the x-axis denotes total training time and the y-axis denotes accuracy. Here, the radius of dashed circles is proportional to the 4th Root of the parameters in the incorporated language models, and the radius of the internal marker is proportional to the number of tunable parameters. Our method with caching achieves a remarkable 72x faster training compared the previous SoTA method, SimTeG, while simultaneously achieving superior performance. Please refer to Table~\ref{tab:train_eff} for more detailed results.}
    \label{fig:dot}
\end{figure}

Large language models (LLMs) such as ChatGPT~\cite{openai2023gpt4}, LLaMA2~\cite{touvron2023llama}, and Falcon~\cite{almazrouei2023falcon} have recently demonstrated significant potential in understanding language, effectively capturing the semantic richness of text attributes. This success has prompted researchers to utilize LLMs on textual graphs for enhanced performance. Some works~\cite{giant,he2023explanations,duan2023simteg} explore the application of Language Models (LMs) on textual graphs to improve node features. These features are subsequently aggregated by GNNs to generate the final representation like Figure~\ref{fig:comparsion}b. 
However, these approaches, as pointed out by many preceeding researchers~\cite{glem,leading,10453398}, may be sub-optimal: the nodes' textual and structural features are encoded separately by the LLMs and GNNs at two different stages. An apparent solution to this issue is the joint training of LMs and GNNs, while this option introduces the efficiency challenges, as memory and time complexity during training and inference may become prohibitively expensive for both institutions and customers.

This work aims to provide an efficient and effective solution to enable the joint modeling of the textual and topological information on textual graphs. We propose an \underline{e}fficie\underline{n}t tunin\underline{g} algor\underline{i}thm for large la\underline{n}guage models on t\underline{e}xtual graphs, named \ours, to address the challenges mentioned above.
As illustrated in Figure~\ref{fig:comparsion}d, during training, we freeze the parameters of LLMs and add a tiny tunable ladder (\emph{G-Ladder}) alongside each layer of LLMs. Within each G-Ladder, we adopt message passing to integrate structural information, thereby enhancing the quality of node representations.
The key idea behind the G-Ladder is that only an extremely small portion of the parameters are updated, resulting in a significant reduction in memory consumption. Furthermore, since the parameter update of \ours does not depend on the gradient computation of the LLMs, which allows us to pre-compute node embeddings, storing them in the cache for subsequent reuse, thereby significantly reducing the time complexity during training, as shown in Figure~\ref{fig:dot}.
In the inference phase, where LLM-induced latency is typically high, we introduce an early exit operator post each ladder to expedite the inference process. This variant, termed \oure, facilitates dynamic early exit of node embedding encoding, accelerating inference and mitigating the risk of overthinking~\cite{zhou2020bert}.
Through extensive experiments, our method demonstrates SoTA performance compared to baselines on textual graphs across various domains. Compared with existing LLM-based approaches, our proposed framework significantly improves training efficiency and reduces inference latency. This suggests a promising direction for combining LLMs with GNNs in textual graph analysis.

Our contributions can be concluded as:

\begin{itemize}
    \item We propose a memory- and time-efficient tuning method for LLMs on textual graphs named \our. 
    The LLMs and GNNs are combined through a \emph{tunable side structure} which significantly reduces the training complexity without impairing the joint model's capacity.
    \item Two additional variants of \ours are proposed to further improve the training and inference: (i) \ours (caching) precomputes node embeddings for training samples, places them into cache, and reuses them during training; and (ii) dynamic early exit dynamically determines whether to exit the feed forward of the LLM layers and returns the prediction.
    \item Extensive experiments demonstrate the effectiveness of our method with the lowest training cost compared to previous methods. Moreover, \ours with caching speeds up training time by 12x, and \oures achieves up to 5x faster inference with a slight maximum performance drop of 1.17\% across 7 datasets.
\end{itemize}

\section{Related Work}
\subsection{LM-based Methods for Textual Graphs}
\begin{figure*}
    \centering
    \includegraphics[scale=0.67]{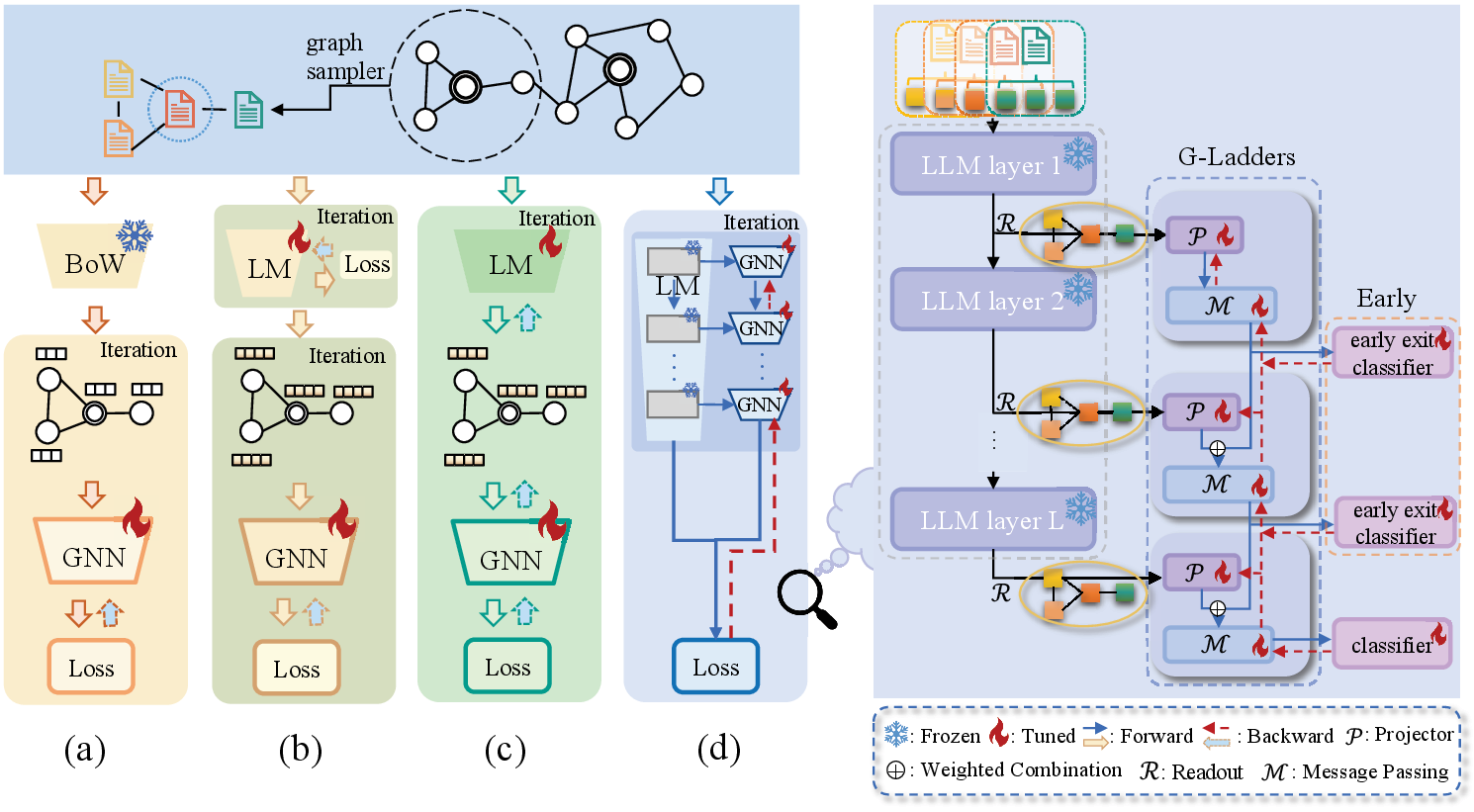}
    \caption{\textbf{Left:} strategies for processing textual graphs on graph tasks: (a) Static shadow embedding method and GNN. (b) Cascading Structure: LMs and GNNs are trained independently. (c) Iterative or Co-training Structure: LMs and GNNs are trained jointly. (d) Side Structure: Frozen LMs combined with tunable GNNs in a side structure. \textbf{Right:} detailed pipeline of \ours, where frozen LLM layers are combined with G-Ladders through a side structure. The dynamic early exit classifier is added after G-Ladders for \oure.}
    \label{fig:comparsion}
\end{figure*}

Representation learning for textual graphs has gained significant attention in the field of graph machine learning. Rather than relying solely on shallow or hand-crafted features (\eg, bag-of-words, skip-gram) as seen in previous works
~\cite{gcn,gat,sage,zhu2023sgl,zhu2023mario,zhu2023graphcontrol}
, LM-based methods harness the potential power of Language Models (LMs) to handle textual graphs. 
These methods combine LMs and GNNs in a cascading (Figure~\ref{fig:comparsion}b) or iterative structure (Figure~\ref{fig:comparsion}c). Specifically, in cascading structure methods, the initial step involves fine-tuning pre-trained language models on downstream tasks. Subsequently, these fine-tuned models are employed to generate text embeddings, serving as the initial node features for GNNs. For example, GIANT~\cite{giant} enhances the meaningfulness of node embeddings by fine-tuning LMs using neighbor information. SimTeG~\cite{duan2023simteg} adopts a two-stage training paradigm: initially fine-tuning LMs through graph-related tasks (\eg, node classification, link prediction), and subsequently training GNNs based on the embeddings generated by LMs. TAPE~\cite{he2023explanations} utilizes LLMs to augment text descriptions and fine-tunes PLMs based on these augmented texts. The subsequent step involves using the fine-tuned PLMs to generate text embeddings, serving as the initial node attributes for GNNs. These methods only model partial information, limiting their ability to learn comprehensive features.

Instead of adopting the cascading structure, GraphFormers \cite{yang2021graphformers}, GLEM \cite{glem}, and LEADING \cite{leading} combine LMs and GNNs in an iterative or co-training structure, training LLMs and GNNs in a co-training way. However, these co-training paradigms face significant scalability issues. Encoding all neighbors by LMs introduces high-cost fine-tuning and inference overhead due to the substantial number of parameters in language models.

To address these non-trivial challenges, we propose to combine LMs and GNNs in a \emph{side structure} as depicted in Figure~\ref{fig:comparsion}d. We introduce a novel method designed to implement a \emph{parameter- and memory-efficient tuning method} for LLMs on textual graphs named \our.

\subsection{Parameter-Efficient Fine-Tuning (PEFT)}
Recently, large language models (LLMs) have achieved remarkable success in NLP domain~\cite{openai2023gpt4,touvron2023llama}. PEFT methods aim to adapt large pre-trained models to various downstream applications without fine-tuning all parameters due to prohibitive costs. For instance, adapter-based methods~\cite{houlsby2019parameter} inject lightweight neural networks (adapters) into Transformer layers and exclusively fine-tune these adapters for model adaptation. LoRA~\cite{lora} assumes a low intrinsic rank in weight changes during model tuning. It proposes optimizing the decomposition of original weight matrices in self-attention modules, multiplying optimized matrices during deployment to obtain the delta of self-attention weights, thus significantly reducing the parameters requiring fine-tuning. Similarly, IA3~\cite{IA3} modifies attention weights for both key and value, as well as the feedforward activation, through element-wise multiplication. Prompt-based tuning~\cite{prompt-tuning,prefix-tuning,pan2023self} prepends trainable soft prompts to the input text while keeping the entire pre-trained models unchanged. While most methods mainly focus on achieving competitive performance through tuning few parameters (parameter-efficient), they often fall short in terms of memory efficiency. Ladder Side-Tuning (LST)~\cite{lst} addresses this issue by training a ladder-side network that separates trainable parameters from the backbone model. Because it does not require backpropagation through the backbone network (LMs), LST significantly reduces memory requirements.

Although PEFT has achieved great success in the Euclidean domain including natural language and computer vision~\cite{cv_prompt,sensitivity_cv_peft,pan2024i3}, how to \emph{effectively} and \emph{efficiently} apply such approaches to non-Euclidean domains like graphs is still under exploration. The main challenge lies in the lack of experience of how to fully utilize structural information to enhance node-level representations.
In this work, we propose an efficient tuning method for LLMs on textual graphs, named \our, which explicitly utilizes structural information and enhances node-level representations through our designed lightweight \textit{G-Ladder}.

\section{Method}
In this section, we will introduce the notations used in this paper. Subsequently, we will present the proposed efficient tuning method for LLMs tailored to textual graphs in Section~\ref{sec:tune}. Finally, efficient inference will be discussed in Section~\ref{sec:inference}.
\subsection{Notations}
Given a textual graph \(G=\{\mathcal{V}, \{t_n\}_{n\in \mathcal{V}},A,Y\}\), where \(\mathcal{V}\) is the node set consisting of \(N\) instances, $t_n \in \mathcal{T}^{Q_n}$ represents a sequential text for its node \(n \in \mathcal{V}\), $\mathcal{T}$ is the tokens dictionary, and $Q_n$ is the sequence length, \(A \in \mathbb{R}^{N\times N}\) denotes the adjacency matrix, and \(Y\) denotes labels for each node. To enhance scalability, a sampling function \(\Gamma(\cdot)\) is applied to a large graph to obtain a set of small subgraphs \(\{G_n\}_{n\in \mathcal{V}}\), where \(G_n\) represents the subgraph for the target node \(n \in \mathcal{V}\). In this study, the focus is on the node classification task of textual graphs. Specifically, given a set of training nodes $\mathcal{V}_{tr}$, a classification model is trained on these nodes and evaluated on the remaining test nodes $\mathcal{V}_{te}$. Formally, given a set of training nodes and their induced subgraphs $\mathcal{G}^{tr} = \{G_n\}_{n\in \mathcal{V}_{tr}}$, the optimal predictor $f_{{\theta}^*}$ is formulated as
\begin{equation}
     f_{{\theta}^*} \in \argmax_{\theta}\mathbb{E}_{G_n \in \mathcal{G}^{tr}}P_{\theta}(\hat{y}_n=y_n \mid G_n),
\end{equation}
where $y_n$ denotes the true label of target node $n \in \mathcal{V}_{tr}$ and $\hat{y}_n$ is the predicted label. It is noteworthy that our method can be applied to other tasks as well, including graph classification and link prediction. For instance, in graph classification, each instance aligns with a graph, analogous to the subgraph of the target node in the node classification task. The exploration of these applications is left for future work.

\subsection{Efficient Tuning for LLM on Textual Graphs}\label{sec:tune}
The efficient tuning of large language models for graph data is under exploration~\cite{duan2023simteg,leading}. However, existing methods are parameter-efficient but not memory-efficient, primarily focusing on fine-tuning small-scale language models like BERT~\cite{kenton2019bert} and DeBERTa~\cite{he2020deberta} for textual graphs. Moreover, these methods encounter challenges in effectively utilizing structural information explicitly during fine-tuning. In this work, we present a parameter-efficient and memory-efficient tuning method for LLMs while also leveraging structural knowledge explicitly, as depicted in Figure~\ref{fig:comparsion} (Right).

Given an input graph $G$, a sampling function $\Gamma(\cdot)$, such as random walk with restart~\cite{gcc,rosa}, will be applied to obtain subgraphs $\mathcal{G}^{tr} = \{G_i\}_{i\in \mathcal{V}_{tr}}$ for target nodes. Each subgraph consists of several nodes, and each node contains a textual description $t_i$. The textual descriptions of nodes within a subgraph will be packed into a batch $\mathcal{B}=\{t_i\}_{i=1}^B$ and fed into LLMs. In each layer of LLM, token-level representations of each node will be calculated:
\begin{equation}
    H^l = \operatorname{LLM\_Layer}^l(H^{l-1}),
\end{equation}
where $\operatorname{LLM\_Layer}^l(\cdot)$ denotes the $l$-th layer of LLM, and $H^l \in \mathbb{R}^{B \times Q \times D}$ means the token-level representations in $i$-th layer. In the first layer, the input $H^0$ is equivalent to $\mathcal{B}$.

\subsubsection{Lightweight \emph{G-Ladders}}
In this work, we focus on node-level tasks, intricately related to the quality of node-level representations. Our goal is to enhance these representations during model tuning. To achieve this, a readout function $\mathcal{R}$, such as mean pooling~\cite{pooling}, is applied to token-level representations:
\begin{equation}
    z_i^{l} = \mathcal{R}(h_{i,1}^l,h_{i,2}^l,...,h_{i,Q}^l),
\end{equation}
where $z_i^{l} \in \mathbb{R}^{1 \times D}$ denotes the representation of node $i \in \mathcal{V}_{tr}$. 
Then these node-level representations are fed into GNN Ladders (\emph{G-Ladders}) to improve the quality of node embeddings through structural information, ensuring accurate comprehension of the node's semantics from a global perspective. The improved node-level representations $\hat{Z}^l$ are derived as
\begin{equation}\label{equ:gladder}
    \hat{Z}^l=\lambda^l\cdot\operatorname{GNN}^l\left(\mathcal{P}^l\left(Z^l\right),A\right) + \left(1-\lambda^l\right)\cdot \hat{Z}^{l-1},
\end{equation}
where $\mathcal{P}^l(\cdot): \mathbb{R}^{B \times D} \rightarrow \mathbb{R}^{B\times K}$ is a projector that maps the node-level representations into low dimensions ($K \ll D$), thereby reducing the subsequent computation in G-Ladders. $\operatorname{GNN}$ is one type of graph neural networks that includes message passing. In this paper, for simplicity, we benchmark three different architectures, including GCN~\cite{gcn}, SAGE~\cite{sage}, and GAT~\cite{gat} and pick the best one~(SAGE) for modeling G-Ladders. Nevertheless, the choice of the GNN architectures does not bring heavy influence to the final performance~(refer to Table 11 in Appendix).
$\lambda^l$ is a learnable coefficient for combining information in the current layer $l$ and the previous layer $l-1$. It is modeled by $\lambda^l=\operatorname{sigmoid}(\frac{\omega^l}{T})$, where $\omega^l$ is a learnable zero-initialized scalar, and $T$ is a temperature set as $0.1$. 

Finally, the representation of the target node $\hat{z}_i^L$ from the final layer's output $\hat{Z}^L$ is fed into a linear classifier $\mathcal{C}$ for classification. The training loss $\mathcal{L}$ is computed using the cross-entropy loss $\operatorname{CE}(\cdot,\cdot)$ between predictions and true labels:
\begin{equation}
    \mathcal{L}=\mathbb{E}_{i \in \mathcal{V}_{tr}}\operatorname{CE}(\hat{y}_i,y_i),\ \text{where}\ \hat{y}_i=\mathcal{C}(\hat{z}_i^L).
\end{equation}

\subsubsection{What Makes \ours An Efficient-Tuning Method?} 
As depicted in Figure~\ref{fig:comparsion} (Right), the efficiency of \ours stems from three aspects. First, similar to existing PEFT methods, the parameters of LLMs are frozen and only lightweight G-Ladders are updated during training, ensuring parameter-efficient. 
Second, our method innovatively integrates \emph{G-Ladders} with LLMs via a \emph{side structure}, eliminating the need for backpropagation through the LLMs. This contrasts with prior methods that insert additional tunable parameters within LLMs, requiring costly backpropagation memory, such as cascading or iterative structures~\cite{giant,glem}.
Last, our side-structure design allows us to precompute and cache node embeddings for reuse, further boosting computational efficiency. In short, our method is both parameter-efficient and memory-efficient, as demonstrated by empirical evidence in Section~\ref{sec:efficiency}.

\subsection{Efficient Inference for LLM on Textual Graphs}\label{sec:inference}
The high latency linked to LLM inference can hinder model deployment in real-world settings.
Various strategies are proposed to speed up model inference, including model pruning \cite{pruning1}, model distillation \cite{sanh2019distilbert}, and dynamic early exit \cite{xin-etal-2020-deebert,smartbert}. 

In this work, for simplicity, we adopt a similar mechanism as dynamic early exit, which can be seamlessly integrated with our method.
Specifically, we add a lightweight single-layer MLP as an early exit classifier $\mathcal{C}^l$ after each \emph{G-Ladder}. 
During the model tuning, these classifiers are directly connected to the downstream task's training objective, \eg, the cross-entropy loss between the true label $y$: 
\begin{equation}
    {C^l}^* \in \argmin_{\mathcal{C}^l}\mathbb{E}_{i \in \mathcal{V}_{tr}}\operatorname{CE}(\mathcal{C}^l(\hat{z}_i^l),y_i).
\end{equation}

Later during inference, early exit will be triggered based on the classifiers' uncertainty measure, such as information entropy~\cite{xin-etal-2020-deebert,smartbert} and confidence~\cite{xie2021elbert}.
However, these criteria heavily depend on datasets, and careful selection is necessary for different datasets. For example, on simpler datasets, a lower value of information entropy may be chosen as the threshold, considering the trade-off between efficiency and performance. In this work, we utilize `patience' criteria~\cite{zhou2020bert} to guide dynamic early exit, allowing us to avoid the need for meticulous design. 
Patience-based criteria imply that if consecutive $p$ early exit classifiers predict the same results, where $p$ represents the number of layers set at 2 across different datasets in our experiments, the model stops inference early and makes a prediction.

Since these early exit classifiers are also included in the side structure, the gradients continue to flow exclusively within the side structure as shown in Figure~\ref{fig:comparsion} (Right), making our method tuning-efficient and inference-efficient.

\section{Experiments}
In this section, we first introduce the datasets used in Section~\ref{sec:datasets}. Then, we will illustrate the baselines and experimental setup in Section~\ref{sec:baselines} and \ref{sec:setup} respectively, and conduct experiments on these datasets to demonstrate the effectiveness of our proposed method in Section~\ref{sec:performance}. Then, we will study the training efficiency of \ours compared to other methods in Section~\ref{sec:efficiency}. Lastly, the analysis of hyper-parameters and an ablation study will be provided. Furthermore, additional experiments (\eg, link prediction) are in Appendix.
\subsection{Datasets}\label{sec:datasets}
In this work, we adopt seven commonly used textual graphs to evaluate our proposed \our: Cora~\cite{collective}, CiteSeer~\cite{giles1998citeseer}, WikiCS~\cite{mernyei2020wiki}, OGBN-ArXiv~\cite{ogb}, ArXiv-2023~\cite{he2023explanations}, OGBN-Products~\cite{ogb} and Ele-Photo~\cite{yan2023comprehensive}. We utilize collected raw text data of these datasets by previous works~\cite{chen2023label,he2023explanations,yan2023comprehensive}. Details of these datasets can be found in Appendix A.
\begin{table}[htbp]
    \centering
    \scalebox{0.92}{
    \begin{tabular}{c|ccc}
    \toprule
        Dataset & \#Nodes & \#Edges   & \#Classes  \\  \midrule
        Cora     &  2,708   & 5,429    & 7         \\
        CiteSeer &  3,186   & 4,277    & 6         \\
        WikiCS   &  11,701  & 215,863  & 10        \\
    OGBN-ArXiv   &  169,343 &1,166,243 & 40        \\
    ArXiv-2023   &  46,198  & 78,543   & 40        \\
OGBN-Products (subset)&54,025&74,420    & 47        \\
    Ele-Photo    &  48,362  &500,928   & 12        \\
\bottomrule
    \end{tabular}}
    \caption{Statistics of textual graphs used in this work.}
    \label{tab:dataset}
\end{table}

\subsection{Baselines}\label{sec:baselines}
To assess the effectiveness of our proposed method, 17 baselines in 5 main categories of approaches are employed. The 5 categories are: (i)~traditional GNN models, (ii)~Graph Transformers, (iii)~LM-Based methods, (iv)~recent works designed for textual graphs, and (v)~PEFT methods.
Briefly, traditional GNN models include \textbf{GCN}~\cite{gcn}, \textbf{SAGE}~\cite{sage}, and \textbf{GAT}~\cite{gat}. Graph Transformers include \textbf{GraphFormers}~\cite{yang2021graphformers} and \textbf{NodeFormer}~\cite{wu2022nodeformer}. Full fine-tuned LM-based methods include \textbf{BERT}~\cite{kenton2019bert}, \textbf{SentenceBERT}~\cite{reimers2019sentence}, and \textbf{DeBERTa}~\cite{he2020deberta}. Recent works for textual graphs include Node Feature Extraction by Self-Supervised Multi-scale Neighborhood Prediction (\textbf{GIANT})~\cite{giant}, Learning on Large-scale Text-attributed Graphs via Variational Inference (\textbf{GLEM})~\cite{glem}, LLM-to-LM Interpreter for Enhanced Text-Attributed Graph Representation Learning (\textbf{TAPE})~\cite{he2023explanations}, and A Frustratingly Simple Approach Improves Textual Graph Learning (\textbf{SimTeG})~\cite{duan2023simteg}. PEFT methods include Low-rank Adaptation of Large Language Models (\textbf{LoRA})~\cite{lora}, IA3~\cite{IA3}, The Power of Scale for Parameter-Efficient Prompt Tuning (\textbf{Prompt Tuning})~\cite{prompt-tuning}, and Ladder Side-Tuning (\textbf{LST})~\cite{lst}. Details of these methods are in Appendix B.

\begin{table*}[htp]
    \centering
    \scalebox{0.94}{
    \begin{tabular}{l|ccccccc}
    \toprule
    Methods         & Cora       & CiteSeer   & WikiCS     & OGBN-ArXiv & ArXiv-2023 & OGBN-Products & Ele-Photo\\ \midrule
    MLP             & 74.32±2.75 & 71.13±1.37 & 68.41±0.65 & 55.54±0.11 & 65.39±0.39 & 56.66±0.10    & 61.21±0.11 \\
    GCN             & 86.90±1.51 & 72.98±1.32 & 76.33±0.81 & 71.51±0.33 & 67.60±0.28 & 69.86±0.14    & 79.00±0.22 \\
    SAGE            & 85.73±0.65 & 73.61±1.90 & 79.56±0.22 & 71.92±0.32 & 69.06±0.24 & 69.75±0.10    & 80.35±0.26 \\
    GAT             & 85.73±0.65 & 74.23±1.78 & 78.21±0.66 & 71.64±0.27 & 67.84±0.23 & 69.57±0.18    & 82.08±0.11 \\
$\text{GraphFormers}^\star$
                    & 80.44±1.89 & 71.28±1.17 & 72.07±0.31 & 67.25±0.22 & 62.87±0.46 & 68.15±0.76    & 75.44±0.56 \\
    NodeFormer      & 88.48±0.33 & 75.74±0.54 & 75.47±0.46 & 69.60±0.08 & 67.44±0.42 & 67.26±0.71    & 77.30±0.06 \\  
 \midrule
    BERT            & 80.15±1.67 & 73.17±1.75 & 78.33±0.43 & 72.78±0.03 & 77.46±0.27 & 76.01±0.14    & 68.88±0.05\\
    SentenceBERT    & 78.82±1.39 & 72.79±1.71 & 77.92±0.07 & 71.42±0.09 & 77.53±0.45 & 75.07±0.13    & 68.74±0.07\\
    DeBERTa         & 77.79±2.26 & 73.13±1.94 & 75.11±1.97 & 72.90±0.05 & 77.25±0.20 & 75.61±0.28    & 70.82±0.08\\ \midrule
    $\text{GIANT}_{(\text{BERT})}$    
                    & 85.52±0.74 & 72.38±0.83 & 75.81±0.26 & 74.26±0.17 & 72.18±0.24 & 74.06±0.42    & 81.27±0.41\\ 
    $\text{GLEM}_{(\text{DeBERTa})}$
                    & 85.60±0.09 & 75.89±0.53 & 78.92±0.19 & 74.69±0.25 & 78.58±0.09 & 73.77±0.12    &76.10±0.23\\
 $\text{TAPE}_{(\text{DeBERTa})}$ 
                    & 88.52±1.12 & $-$        & $-$        & 74.65±0.10 & 79.23±0.52 & 79.76±0.11    & $-$ \\
 $\text{SimTeG}_{(\text{e5-large})}$
                    & 88.04±1.36 & 77.22±1.43 & 79.07±0.65 & 75.29±0.23 & \underline{79.51±0.48} & 74.51±1.49    & 83.07±0.20\\ 
    \midrule
LoRA (\logo)        
                    & 79.95±0.44 & 73.61±1.89 & 78.91±1.26 & 74.94±0.03  & 78.85±0.21 & 75.50±0.05   & 73.25±0.03 \\
IA3 (\logo)  
              & 76.43±1.29 & 71.07±1.24 & 70.08±1.26 & 71.87±0.03  & 78.14±0.30 & 75.82±0.10   & 69.27±0.37 \\
Prompt Tuning (\logo)   
              & 73.73±2.05 & 69.62±2.14 & 67.14±1.50 & 71.34±0.58  & 74.78±0.70 & 74.50±0.99   & 62.84±0.27 \\
LST (\logo)          
              & 77.60±0.76 & 75.05±1.36 & 77.59±0.70 & 73.68±0.90  & 77.82±0.37 & 76.10±0.79   & 68.93±0.21\\
\rowcolor{Gray}
\our (\logo)        
              & \textbf{91.48±0.32} & \textbf{78.46±0.49} & \textbf{81.56±0.97} & \textbf{76.02±0.29} & \textbf{79.76±0.14} & \textbf{80.05±0.45}    & \textbf{83.75±0.08} \\ 
\rowcolor{Gray}
\oure (\logo)   
              & \underline{90.41±0.52} & \underline{78.34±0.75} & \underline{81.23±0.78} & \underline{75.66±0.54} & 79.29±0.18 & \underline{79.78±0.62} & \underline{83.13±0.44}\\ \midrule
    \end{tabular}}
    \caption{Experimental results of node classification. \logo denotes LLaMA2-7B model. \oures means that use dynamic early exit to accelerate model inference. $\star$ denotes our adaption of their official codes to our tasks. We use \textbf{boldface} and \underline{underlining} to denote the best and the second-best performance, respectively.
    }
    \label{tab:main}
\end{table*}

\subsection{Experimental Setup}\label{sec:setup}
For traditional GNN methods, we utilize grid search to obtain optimal results. For LM-based methods, \ie, BERT, SentenceBERT, DeBERTa, we fine-tune all parameters of these models on training nodes. Methods tailored for textual graphs are implemented using their official codes and reproduced under our settings. The GNNs utilized in these methods are selected from GCN, SAGE, and GAT, choosing the most effective one. Notably, GraphFormers is originally designed for link prediction task, $\text{GraphFormers}^\star$ refers to our adaptation of their official codes\footnote{https://github.com/microsoft/GraphFormers} to the node classification tasks. The hyperparameters of the baselines can be found in Appendix C.
Regarding our method, \ours can be applied to any LLMs. In the main content, we applied \ours to the widely used open LLM named LLaMA2-7B~\cite{touvron2023llama} to show the effectiveness. \textbf{Experimental results for other language models, like e5-large~\cite{e5-large}, can be found in Appendix D.} We report the mean accuracy with a standard deviation across five different random seeds.

\subsection{Performance Analysis}\label{sec:performance}
From Table~\ref{tab:main}, we can draw the following conclusions:

Firstly, static shallow embedding methods combined with GNNs (\ie, GCN, SAGE, GAT) perform inferiorly compared to recent methods that combine LMs with GNNs. This indicates that static shallow embedding methods may struggle to capture context-aware information and complex semantic relationships, limiting their ability to fully exploit the richness of text attributes, thereby achieving suboptimal results. For instance, on OGBN-ArXiv and OGBN-Products datasets, LM+GNN methods (\ie, SimTeG, GLEM, GIANT) significantly outperform GNNs with shallow embedding methods, exhibiting an absolute performance gap of around 5\%.

Secondly, pure LM methods (\ie, BERT, SentenceBERT, DeBERTa) perform inferiorly to LM+GNN methods on textual graphs. This demonstrates that, compared to pure LM methods which neglect intrinsic graph structure, combining LMs with GNNs can generate more semantic and structure-aware node embeddings. For instance, on the Ele-Photo dataset, GIANT achieves 81\% accuracy, outperforming its base LM model BERT by approximately 13\%.

Lastly, our method is superior to current LM+GNN methods. Specifically, \ours further outperforms the previous SoTA method SimTeG, achieving an absolute improvement of over 2\% on the Cora dataset and an absolute 3\% improvement on the WikiCS dataset. Besides, our method also shows significant improvements over all the other PEFT methods (\ie, LoRA, IA3, Prompt Tuning, LST), showcasing the superiority of \ours in fine-tuning LLMs on textual graphs. Additionally, \oure, which combines dynamic early exiting, achieves comparable performance with \ours while improving the efficiency of model inference, as further discussed in Section~\ref{sec:efficiency}.

These statistics show the effectiveness of our method for generating context-aware and complex semantic nodes embeddings on textual graphs.

\subsection{Efficiency Analysis}\label{sec:efficiency}
\subsubsection{Training Efficiency}

In this section, we assess the training efficiency of several baselines and \our. The additional trainable parameters in our method are integrated with a frozen LLM through a side structure. This allows us to precompute node embeddings using frozen LLMs and store them in the cache for reuse, consequently enhancing training efficiency. We also present the training efficiency of our method with caching.

\begin{table}[htp]
    \centering
    \scalebox{0.88}{
    \begin{tabular}{l|c|c|c}
    \toprule
     Methods    &  Update Param. & Memory (GB) & Total Time \\ \midrule
     
     $\text{GIANT}_{\text{(BERT)}}$       &                114,535,896&        7.5&   29h 44m   \\ 
     $\text{GLEM}_{\text{(DeBERTa)}}$        &                138,731,759&        6.6&      46h 27m\\ 
     $\text{SimTeG}_{\text{(e5-large)}}$      &   2,013,328&  3.7&      25h 22m\\ \midrule
     LoRA (\logo)       & 4,358,144      & 44.8   &10h 18m \\ 
     IA3 (\logo)        & 425,984        & 28.5  & 9h 55m\\ 
     Prompt. (\logo)   & 245,760        & 27.8  & 9h 48m  \\ \midrule
     \rowcolor{Gray}
     \our (\logo)       & 3,909,032       & 14.9   & 4h 23m  \\ 
     \rowcolor{Gray}
      \quad w/ caching& 3,909,032 & 3.3    & 21m     \\ 
     \bottomrule
    \end{tabular}}
    \caption{The efficiency analysis of training different methods on OGBN-ArXiv dataset. The batch size is set as 1 for tuning LMs, and the total training time is reported on the CPU of a 48-core Intel(R) Xeon(R) @ 2.50GHz and GPUs of 6 NVIDIA GeForce RTX 3090.}
    \label{tab:train_eff}
\end{table}

In Table~\ref{tab:train_eff}, GIANT and GLEM employ full fine-tuning of language models with the largest trainable parameters. In contrast, SimTeG utilizes LoRA to fine-tune e5-large and subsequently trains GNNs. These methods consume low memory because they employ small language models. However, scaling them up to large language models proves challenging due to their high training costs.
Furthermore, it is evident that previous PEFT methods demand substantial time and memory resources for tuning LLaMA2-7B. In contrast, \ours achieves the lowest time and memory costs, and the inclusion of caching in \ours significantly reduces computation expenses, achieving 12x faster training compared to without caching (21m vs 4h 23m).

\subsubsection{Inference Efficiency}
\begin{figure}
    \centering
    \includegraphics[scale=0.68]{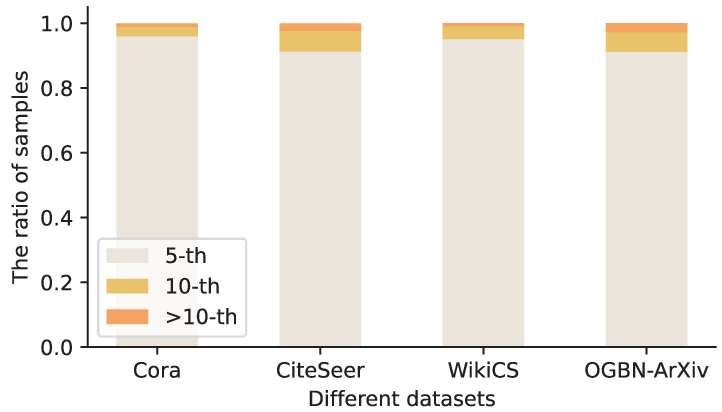}
    \caption{The statistics of samples early exit at each layer.}
    \label{fig:bar}
\end{figure}

For inference, \ours goes through all layers in LLMs, which may be impractical in some realistic scenarios. Therefore,  we incorporate dynamic early exit seamlessly into our method, named \oure. \oures dynamically exit based on the complexity of samples to save inference time. Table~\ref{tab:inference_time} records the inference time for running all test samples using our method (with or without dynamic exit). Figure~\ref{fig:bar} depicts the early exit ratios across different layers.

\begin{table}[htp]
    \centering
    \scalebox{0.95}{
    \begin{tabular}{l|c|c|c}
    \toprule
Methods&Cora & CiteSeer & WikiCS\\ \midrule
$\text{GIANT}_{(\text{BERT})}$  &      41s&      47s&          2m 58s\\
$\text{GLEM}_{(\text{DeBERTa})}$   &    1m 8s  &         1m 19s&     5m 27s     \\
$\text{SimTeG}_{(\text{e5-large})}$ &      1m 21s&         1m 35s&          5m 53s\\ \midrule
\rowcolor{Gray}
\ours(\logo)    &1m 30s & 1m 54s & 15m 55s  \\
\rowcolor{Gray}
\oures(\logo)   &15s & 20s&  2m 38s\\ 
     \bottomrule
    \end{tabular}}
    \caption{Inference time of \our and baselines across datasets.}
    \label{tab:inference_time}
\end{table}

In Table~\ref{tab:inference_time}, the speed advantage of \oures over \our is clearly evident, primarily due to reduced intensive computation within LLM layers. Notably, \oures attains 5x faster inference speed, mainly attributed to the fact that over 90\% of samples exit early in the 5th layer as depicted in Figure~\ref{fig:bar}. Additionally, \oures is faster than other textual graph baselines while achieving better performance.

\subsection{Sensitivity Analysis}\label{sec:sensitivity}
\paragraph{The number of G-Ladders.} G-Ladder can be incorporated alongside each LLM layer. In our investigation, we examine varying numbers of G-Ladders. For simplicity, we analyze three strategies:
(1) Adding a G-Ladder every 10 layers (\ie, $0, 10, 20, 32$ where the 0-th layer represents the word embedding layer, and the 32-nd layer is the last layer used in all strategies).
(2) Adding a G-Ladder every 5 layers (\ie, $0, 5, ..., 25, 32$).
(3) Adding a G-Ladder every 2 layers (\ie, $0, 2, ..., 30, 32$).

Table~\ref{tab:num_ladders} presents the results. The first strategy consistently performs worse than others, likely due to having the fewest learnable parameters. Although the last strategy involves the most learnable parameters, it falls short of the second strategy on small datasets (\eg, Cora), suggesting that a moderate number of parameters is optimal. Throughout our experiments, we adopt the second strategy due to its balanced consideration of effectiveness and efficiency.

\begin{table}[htp]
    \centering
    \scalebox{0.98}{
    \begin{tabular}{l|c|c|c}
    \toprule
Inserted Layers&Cora & CiteSeer & WikiCS\\ \midrule
      $0, 10, 20, 32$    & 90.22±1.05 &  78.31±0.68& 81.20±0.86\\
      $0, 5\ \ , ..., 25, 32$ & 91.48±0.32 & 78.46±0.49 & 81.56±0.97 \\
      $0, 2\ \ , ..., 30, 32$ & 90.77±0.56& 78.71±0.77 & 81.90±0.76\\
        \bottomrule
    \end{tabular}}
    \caption{Sensitivity analysis of the number of G-Ladders. We report the mean accuracy with a standard deviation across 5 different random seeds.}
    \label{tab:num_ladders}
\end{table}

\paragraph{The patience in dynamic early exit.}
To accelerate inference, we adopt patience-based early dynamic exiting in our work. In this section, experiments are conducted with varying patience values: 2, 3, 4, and 7. The patience of 2 means terminating in advance and returning results if two consecutive layers produce identical results. Notably, the patience of 7 signifies the absence of dynamic early exiting.

\begin{figure}
    \centering
    \includegraphics[scale=0.62]{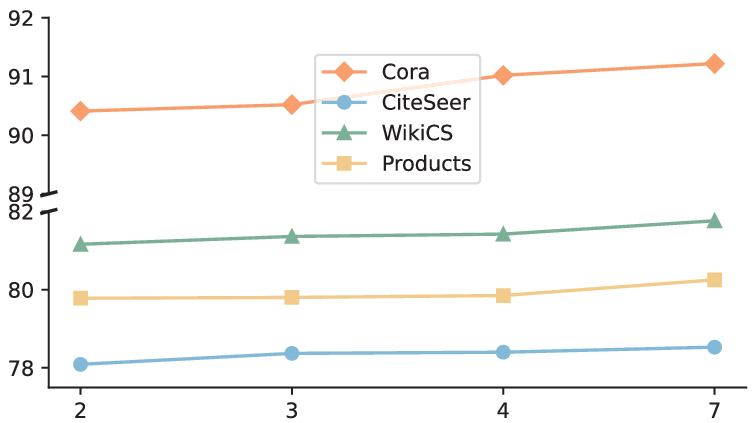}
    \caption{Sensitivity analysis of patience $p$ in dynamic early exit.}
    \label{fig:patience}
\end{figure}

Results from Figure~\ref{fig:patience} illustrate the trade-off between performance and efficiency. In our experiments, we set the patience to 2 for most datasets, achieving comparable performance while significantly reducing inference time, denoted as \oure.

\subsection{Ablation Study}

In this section, we investigate the impact of components in our method. Specifically, `constant $\lambda$ (0.5)' indicates setting $\lambda$ as a constant value (\ie, 0.5) in Equation~\ref{equ:gladder} rather than a learnable coefficient. `w/o struct. info.' denotes the removal of message passing in G-Ladders.

Table~\ref{tab:ablation} highlights that structural information is crucial for enhancing the quality of node representations, showing around a 14\% absolute improvement on Cora and WikiCS datasets. Additionally, a learnable coefficient $\lambda$ outperforms a constant value because the coefficient can adapt to the characteristics of datasets.

\begin{table}[htp]
    \centering
    \scalebox{0.92}{
    \begin{tabular}{l|c|c|c}
    \toprule
               &Cora & CiteSeer & WikiCS\\ \midrule
      \ours    & 91.48±0.32 & 78.46±0.49 & 81.56±0.97 \\
\quad constant $\lambda$ (0.5) & 90.07±0.91 & 78.06±0.36& 80.66±0.66 \\
        \quad w/o struct. info. & 76.94±1.98 &  69.97±3.46& 68.44±2.75\\
        \bottomrule
    \end{tabular}}
    \caption{Experimental results of ablation study. We report the mean accuracy with a standard deviation across 5 different random seeds.}
    \label{tab:ablation}
\end{table}
\section{Conclusion}
In this paper, we present \ours, an efficient and effective framework for incorporating large language models in textual graphs. 
Our proposed approach introduces a lightweight, tunable GNN-based side structure~(G-Ladder) alongside each layer of the LLM, to explicitly model the structural information of the textual graphs. 
The key insight is that the parameter update of \ours does not depend on the gradient computation of the LLMs, resulting in exceptionally efficient training compared to concurrent counterparts. 
Building upon this, we introduce two variants with caching and dynamic early exit to further enhance training and inference speed.
Empirical studies demonstrate that our \ours outperforms the state-of-the-art approaches across multiple realistic textual graph datasets, in terms of performance, training efficiency, and inference efficiency.
\section*{Acknowledgments}
This work was supported by the Key Research and Development Projects in Zhejiang Province (No. 2024C01106), NSFC (No. 62272411), the Tencent WeChat Rhino-Bird Special Research Program (Tencent WXG-FR-2023-10), the National Key Research and Development Project of China (2018AAA0101900), and Research funding from FinVolution Group.
\section*{Contribution Statement}
The paper's methodology and experimental design were collaboratively conceived by all contributing authors. Both \textbf{Yun Zhu} and \textbf{Yaoke Wang} made significant and equal contributions to this work, thereby meriting the designation of co-first authors. \textbf{Siliang Tang} holds the role of corresponding author for this publication.

\bibliographystyle{named}
\bibliography{simple}

\begin{thebibliography}{}

\bibitem[\protect\citeauthoryear{Almazrouei \bgroup \em et al.\egroup }{2023}]{almazrouei2023falcon}
Ebtesam Almazrouei, Hamza Alobeidli, Abdulaziz Alshamsi, Alessandro Cappelli, Ruxandra Cojocaru, Merouane Debbah, Etienne Goffinet, Daniel Heslow, Julien Launay, Quentin Malartic, et~al.
\newblock Falcon-40b: an open large language model with state-of-the-art performance.
\newblock {\em Findings of the Association for Computational Linguistics: ACL}, 2023.

\bibitem[\protect\citeauthoryear{Chen \bgroup \em et al.\egroup }{2023}]{chen2023label}
Zhikai Chen, Haitao Mao, Hongzhi Wen, Haoyu Han, Wei Jin, Haiyang Zhang, Hui Liu, and Jiliang Tang.
\newblock Label-free node classification on graphs with large language models (llms).
\newblock {\em arXiv preprint arXiv:2310.04668}, 2023.

\bibitem[\protect\citeauthoryear{Chiang \bgroup \em et al.\egroup }{2019}]{chiang2019cluster}
Wei-Lin Chiang, Xuanqing Liu, Si~Si, Yang Li, Samy Bengio, and Cho-Jui Hsieh.
\newblock Cluster-gcn: An efficient algorithm for training deep and large graph convolutional networks.
\newblock In {\em Proc. of KDD}, 2019.

\bibitem[\protect\citeauthoryear{Chien \bgroup \em et al.\egroup }{2021}]{giant}
Eli Chien, Wei-Cheng Chang, Cho-Jui Hsieh, Hsiang-Fu Yu, Jiong Zhang, Olgica Milenkovic, and Inderjit~S Dhillon.
\newblock Node feature extraction by self-supervised multi-scale neighborhood prediction.
\newblock In {\em Proc. of ICLR}, 2021.

\bibitem[\protect\citeauthoryear{Duan \bgroup \em et al.\egroup }{2023}]{duan2023simteg}
Keyu Duan, Qian Liu, Tat-Seng Chua, Shuicheng Yan, Wei~Tsang Ooi, Qizhe Xie, and Junxian He.
\newblock Simteg: A frustratingly simple approach improves textual graph learning.
\newblock {\em arXiv preprint arXiv:2308.02565}, 2023.

\bibitem[\protect\citeauthoryear{Giles \bgroup \em et al.\egroup }{1998}]{giles1998citeseer}
C~Lee Giles, Kurt~D Bollacker, and Steve Lawrence.
\newblock Citeseer: An automatic citation indexing system.
\newblock In {\em Proceedings of the third ACM conference on Digital libraries}, 1998.

\bibitem[\protect\citeauthoryear{Gordon \bgroup \em et al.\egroup }{2020}]{pruning1}
Mitchell~A. Gordon, Kevin Duh, and Nicholas Andrews.
\newblock Compressing {BERT:} studying the effects of weight pruning on transfer learning.
\newblock {\em CoRR}, 2020.

\bibitem[\protect\citeauthoryear{Hamilton \bgroup \em et al.\egroup }{2017}]{sage}
Will Hamilton, Zhitao Ying, and Jure Leskovec.
\newblock Inductive representation learning on large graphs.
\newblock {\em Proc. of NeurIPS}, 2017.

\bibitem[\protect\citeauthoryear{He \bgroup \em et al.\egroup }{2020}]{he2020deberta}
Pengcheng He, Xiaodong Liu, Jianfeng Gao, and Weizhu Chen.
\newblock Deberta: Decoding-enhanced bert with disentangled attention.
\newblock In {\em Proc. of ICLR}, 2020.

\bibitem[\protect\citeauthoryear{He \bgroup \em et al.\egroup }{2023a}]{sensitivity_cv_peft}
Haoyu He, Jianfei Cai, Jing Zhang, Dacheng Tao, and Bohan Zhuang.
\newblock Sensitivity-aware visual parameter-efficient fine-tuning.
\newblock In {\em Proc. of ICCV}, 2023.

\bibitem[\protect\citeauthoryear{He \bgroup \em et al.\egroup }{2023b}]{he2023explanations}
Xiaoxin He, Xavier Bresson, Thomas Laurent, and Bryan Hooi.
\newblock Explanations as features: Llm-based features for text-attributed graphs.
\newblock {\em arXiv preprint arXiv:2305.19523}, 2023.

\bibitem[\protect\citeauthoryear{Houlsby \bgroup \em et al.\egroup }{2019}]{houlsby2019parameter}
Neil Houlsby, Andrei Giurgiu, Stanislaw Jastrzebski, Bruna Morrone, Quentin De~Laroussilhe, Andrea Gesmundo, Mona Attariyan, and Sylvain Gelly.
\newblock Parameter-efficient transfer learning for nlp.
\newblock In {\em Proc. of ICML}, 2019.

\bibitem[\protect\citeauthoryear{Hu \bgroup \em et al.\egroup }{2020}]{ogb}
Weihua Hu, Matthias Fey, Marinka Zitnik, Yuxiao Dong, Hongyu Ren, Bowen Liu, Michele Catasta, and Jure Leskovec.
\newblock Open graph benchmark: Datasets for machine learning on graphs.
\newblock {\em Proc. of NeurIPS}, 2020.

\bibitem[\protect\citeauthoryear{Hu \bgroup \em et al.\egroup }{2021}]{lora}
Edward~J Hu, Phillip Wallis, Zeyuan Allen-Zhu, Yuanzhi Li, Shean Wang, Lu~Wang, Weizhu Chen, et~al.
\newblock Lora: Low-rank adaptation of large language models.
\newblock In {\em Proc. of ICLR}, 2021.

\bibitem[\protect\citeauthoryear{Hu \bgroup \em et al.\egroup }{2023}]{smartbert}
Boren Hu, Yun Zhu, Jiacheng Li, and Siliang Tang.
\newblock Smartbert: A promotion of dynamic early exiting mechanism for accelerating bert inference.
\newblock In {\em Proc. of IJCAI}, 2023.

\bibitem[\protect\citeauthoryear{Jang \bgroup \em et al.\egroup }{2016}]{gumbel}
Eric Jang, Shixiang Gu, and Ben Poole.
\newblock Categorical reparameterization with gumbel-softmax.
\newblock In {\em Proc. of ICLR}, 2016.

\bibitem[\protect\citeauthoryear{Jia \bgroup \em et al.\egroup }{2022}]{cv_prompt}
Menglin Jia, Luming Tang, Bor-Chun Chen, Claire Cardie, Serge Belongie, Bharath Hariharan, and Ser-Nam Lim.
\newblock Visual prompt tuning.
\newblock In {\em Proc. of ECCV}, 2022.

\bibitem[\protect\citeauthoryear{Kenton and Toutanova}{2019}]{kenton2019bert}
Jacob Devlin Ming-Wei~Chang Kenton and Lee~Kristina Toutanova.
\newblock Bert: Pre-training of deep bidirectional transformers for language understanding.
\newblock In {\em Proc. of AACL}, 2019.

\bibitem[\protect\citeauthoryear{Kipf and Welling}{2017}]{gcn}
Thomas~N. Kipf and Max Welling.
\newblock Semi-supervised classification with graph convolutional networks.
\newblock In {\em Proc. of ICLR}, 2017.

\bibitem[\protect\citeauthoryear{Lester \bgroup \em et al.\egroup }{2021}]{prompt-tuning}
Brian Lester, Rami Al-Rfou, and Noah Constant.
\newblock The power of scale for parameter-efficient prompt tuning.
\newblock In {\em Proc. of EMNLP}, 2021.

\bibitem[\protect\citeauthoryear{Li and Liang}{2021}]{prefix-tuning}
Xiang~Lisa Li and Percy Liang.
\newblock Prefix-tuning: Optimizing continuous prompts for generation.
\newblock In {\em Proc. of ACL}, 2021.

\bibitem[\protect\citeauthoryear{Liu \bgroup \em et al.\egroup }{2022}]{IA3}
Haokun Liu, Derek Tam, Mohammed Muqeeth, Jay Mohta, Tenghao Huang, Mohit Bansal, and Colin~A Raffel.
\newblock Few-shot parameter-efficient fine-tuning is better and cheaper than in-context learning.
\newblock {\em Proc. of NeurIPS}, 2022.

\bibitem[\protect\citeauthoryear{Mernyei and Cangea}{2020}]{mernyei2020wiki}
P{\'e}ter Mernyei and C{\u{a}}t{\u{a}}lina Cangea.
\newblock Wiki-cs: A wikipedia-based benchmark for graph neural networks.
\newblock {\em arXiv preprint arXiv:2007.02901}, 2020.

\bibitem[\protect\citeauthoryear{Mesquita \bgroup \em et al.\egroup }{2020}]{pooling}
Diego Mesquita, Amauri Souza, and Samuel Kaski.
\newblock Rethinking pooling in graph neural networks.
\newblock {\em Proc. of NeurIPS}, 2020.

\bibitem[\protect\citeauthoryear{Mikolov \bgroup \em et al.\egroup }{2013}]{mikolov2013distributed}
Tomas Mikolov, Ilya Sutskever, Kai Chen, Greg~S Corrado, and Jeff Dean.
\newblock Distributed representations of words and phrases and their compositionality.
\newblock {\em Proc. of NeurIPS}, 2013.

\bibitem[\protect\citeauthoryear{Ni \bgroup \em et al.\egroup }{2019}]{ni2019justifying}
Jianmo Ni, Jiacheng Li, and Julian McAuley.
\newblock Justifying recommendations using distantly-labeled reviews and fine-grained aspects.
\newblock In {\em Proc. of EMNLP}, 2019.

\bibitem[\protect\citeauthoryear{OpenAI}{2023}]{openai2023gpt4}
OpenAI.
\newblock Gpt-4 technical report, 2023.

\bibitem[\protect\citeauthoryear{Pan \bgroup \em et al.\egroup }{2023}]{pan2023self}
Kaihang Pan, Juncheng Li, Hongye Song, Jun Lin, Xiaozhong Liu, and Siliang Tang.
\newblock Self-supervised meta-prompt learning with meta-gradient regularization for few-shot generalization.
\newblock {\em arXiv preprint arXiv:2303.12314}, 2023.

\bibitem[\protect\citeauthoryear{Pan \bgroup \em et al.\egroup }{2024a}]{pan2024i3}
Kaihang Pan, Juncheng Li, Wenjie Wang, Hao Fei, Hongye Song, Wei Ji, Jun Lin, Xiaozhong Liu, Tat-Seng Chua, and Siliang Tang.
\newblock I3: Intent-introspective retrieval conditioned on instructions, 2024.

\bibitem[\protect\citeauthoryear{Pan \bgroup \em et al.\egroup }{2024b}]{10453398}
Shirui Pan, Yizhen Zheng, and Yixin Liu.
\newblock Integrating graphs with large language models: Methods and prospects.
\newblock {\em IEEE Intelligent Systems}, 39(1):64--68, 2024.

\bibitem[\protect\citeauthoryear{Qiu \bgroup \em et al.\egroup }{2020}]{gcc}
Jiezhong Qiu, Qibin Chen, Yuxiao Dong, Jing Zhang, Hongxia Yang, Ming Ding, Kuansan Wang, and Jie Tang.
\newblock Gcc: Graph contrastive coding for graph neural network pre-training.
\newblock In {\em Proc. of KDD}, 2020.

\bibitem[\protect\citeauthoryear{Reimers and Gurevych}{2019}]{reimers2019sentence}
Nils Reimers and Iryna Gurevych.
\newblock Sentence-bert: Sentence embeddings using siamese bert-networks.
\newblock In {\em Proc. of EMNLP}, 2019.

\bibitem[\protect\citeauthoryear{Sanh \bgroup \em et al.\egroup }{2019}]{sanh2019distilbert}
Victor Sanh, Lysandre Debut, Julien Chaumond, and Thomas Wolf.
\newblock Distilbert, a distilled version of bert: smaller, faster, cheaper and lighter.
\newblock {\em arXiv preprint arXiv:1910.01108}, 2019.

\bibitem[\protect\citeauthoryear{Sen \bgroup \em et al.\egroup }{2008}]{collective}
Prithviraj Sen, Galileo Namata, Mustafa Bilgic, Lise Getoor, Brian Galligher, and Tina Eliassi-Rad.
\newblock Collective classification in network data.
\newblock {\em AI magazine}, 2008.

\bibitem[\protect\citeauthoryear{Sung \bgroup \em et al.\egroup }{2022}]{lst}
Yi-Lin Sung, Jaemin Cho, and Mohit Bansal.
\newblock Lst: Ladder side-tuning for parameter and memory efficient transfer learning.
\newblock {\em Proc. of NeurIPS}, 2022.

\bibitem[\protect\citeauthoryear{Touvron \bgroup \em et al.\egroup }{2023}]{touvron2023llama}
Hugo Touvron, Louis Martin, Kevin Stone, Peter Albert, Amjad Almahairi, Yasmine Babaei, Nikolay Bashlykov, Soumya Batra, Prajjwal Bhargava, Shruti Bhosale, et~al.
\newblock Llama 2: Open foundation and fine-tuned chat models.
\newblock {\em arXiv preprint arXiv:2307.09288}, 2023.

\bibitem[\protect\citeauthoryear{Veli{\v{c}}kovi{\'c} \bgroup \em et al.\egroup }{2018}]{gat}
Petar Veli{\v{c}}kovi{\'c}, Guillem Cucurull, Arantxa Casanova, Adriana Romero, Pietro Li{\`o}, and Yoshua Bengio.
\newblock Graph attention networks.
\newblock In {\em Proc. of ICLR}, 2018.

\bibitem[\protect\citeauthoryear{Wang \bgroup \em et al.\egroup }{2020}]{wang2020microsoft}
Kuansan Wang, Zhihong Shen, Chiyuan Huang, Chieh-Han Wu, Yuxiao Dong, and Anshul Kanakia.
\newblock Microsoft academic graph: When experts are not enough.
\newblock {\em Quantitative Science Studies}, 2020.

\bibitem[\protect\citeauthoryear{Wang \bgroup \em et al.\egroup }{2022}]{e5-large}
Liang Wang, Nan Yang, Xiaolong Huang, Binxing Jiao, Linjun Yang, Daxin Jiang, Rangan Majumder, and Furu Wei.
\newblock Text embeddings by weakly-supervised contrastive pre-training.
\newblock {\em arXiv preprint arXiv:2212.03533}, 2022.

\bibitem[\protect\citeauthoryear{Wu \bgroup \em et al.\egroup }{2022}]{wu2022nodeformer}
Qitian Wu, Wentao Zhao, Zenan Li, David~P Wipf, and Junchi Yan.
\newblock Nodeformer: A scalable graph structure learning transformer for node classification.
\newblock {\em Proc. of NeurIPS}, 2022.

\bibitem[\protect\citeauthoryear{Xie \bgroup \em et al.\egroup }{2021}]{xie2021elbert}
Keli Xie, Siyuan Lu, Meiqi Wang, and Zhongfeng Wang.
\newblock Elbert: Fast albert with confidence-window based early exit.
\newblock In {\em Proc. of ICASSP}, 2021.

\bibitem[\protect\citeauthoryear{Xin \bgroup \em et al.\egroup }{2020}]{xin-etal-2020-deebert}
Ji~Xin, Raphael Tang, Jaejun Lee, Yaoliang Yu, and Jimmy Lin.
\newblock {D}ee{BERT}: Dynamic early exiting for accelerating {BERT} inference.
\newblock In {\em Proc. of ACL}, 2020.

\bibitem[\protect\citeauthoryear{Xue \bgroup \em et al.\egroup }{2023}]{leading}
Rui Xue, Xipeng Shen, Ruozhou Yu, and Xiaorui Liu.
\newblock Efficient large language models fine-tuning on graphs.
\newblock {\em arXiv preprint arXiv:2312.04737}, 2023.

\bibitem[\protect\citeauthoryear{Yan \bgroup \em et al.\egroup }{2023}]{yan2023comprehensive}
Hao Yan, Chaozhuo Li, Ruosong Long, Chao Yan, Jianan Zhao, Wenwen Zhuang, Jun Yin, Peiyan Zhang, Weihao Han, Hao Sun, et~al.
\newblock A comprehensive study on text-attributed graphs: Benchmarking and rethinking.
\newblock In {\em Proc. of NeurIPS}, 2023.

\bibitem[\protect\citeauthoryear{Yang \bgroup \em et al.\egroup }{2021}]{yang2021graphformers}
Junhan Yang, Zheng Liu, Shitao Xiao, Chaozhuo Li, Defu Lian, Sanjay Agrawal, Amit Singh, Guangzhong Sun, and Xing Xie.
\newblock Graphformers: Gnn-nested transformers for representation learning on textual graph.
\newblock {\em Proc. of NeurIPS}, 2021.

\bibitem[\protect\citeauthoryear{Zhang \bgroup \em et al.\egroup }{2010}]{zhang2010understanding}
Yin Zhang, Rong Jin, and Zhi-Hua Zhou.
\newblock Understanding bag-of-words model: a statistical framework.
\newblock {\em International journal of machine learning and cybernetics}, 2010.

\bibitem[\protect\citeauthoryear{Zhao \bgroup \em et al.\egroup }{2022}]{glem}
Jianan Zhao, Meng Qu, Chaozhuo Li, Hao Yan, Qian Liu, Rui Li, Xing Xie, and Jian Tang.
\newblock Learning on large-scale text-attributed graphs via variational inference.
\newblock In {\em Proc. of ICLR}, 2022.

\bibitem[\protect\citeauthoryear{Zhou \bgroup \em et al.\egroup }{2020}]{zhou2020bert}
Wangchunshu Zhou, Canwen Xu, Tao Ge, Julian McAuley, Ke~Xu, and Furu Wei.
\newblock Bert loses patience: Fast and robust inference with early exit.
\newblock {\em Proc. of NeurIPS}, 2020.

\bibitem[\protect\citeauthoryear{Zhu \bgroup \em et al.\egroup }{2022}]{rosa}
Yun Zhu, Jianhao Guo, Fei Wu, and Siliang Tang.
\newblock Rosa: A robust self-aligned framework for node-node graph contrastive learning.
\newblock In {\em Proc. of IJCAI}, 2022.

\bibitem[\protect\citeauthoryear{Zhu \bgroup \em et al.\egroup }{2023a}]{zhu2023sgl}
Yun Zhu, Jianhao Guo, and Siliang Tang.
\newblock Sgl-pt: A strong graph learner with graph prompt tuning.
\newblock {\em arXiv preprint arXiv:2302.12449}, 2023.

\bibitem[\protect\citeauthoryear{Zhu \bgroup \em et al.\egroup }{2023b}]{zhu2023mario}
Yun Zhu, Haizhou Shi, Zhenshuo Zhang, and Siliang Tang.
\newblock Mario: Model agnostic recipe for improving ood generalization of graph contrastive learning.
\newblock {\em arXiv preprint arXiv:2307.13055}, 2023.

\bibitem[\protect\citeauthoryear{Zhu \bgroup \em et al.\egroup }{2023c}]{zhu2023graphcontrol}
Yun Zhu, Yaoke Wang, Haizhou Shi, Zhenshuo Zhang, and Siliang Tang.
\newblock Graphcontrol: Adding conditional control to universal graph pre-trained models for graph domain transfer learning, 2023.

\end{thebibliography}

\appendix

\section{Datasets}\label{app:dataset}
This section provides a detailed introduction to the datasets used in the main content:

\textbf{Cora}~\cite{collective} dataset consists of 2,708 scientific publications categorized into seven classes: case-based, genetic algorithms, neural networks, probabilistic methods, reinforcement learning, rule learning, and theory. Each paper in the citation network cites or is cited by at least one other paper, resulting in a total of 5,429 edges. We use the collected dataset\footnote{https://github.com/XiaoxinHe/TAPE} with raw texts provided by TAPE~\cite{he2023explanations}.

\textbf{CiteSeer}~\cite{giles1998citeseer} dataset consists of 3,186 scientific publications categorized into six classes: Agents, Machine Learning, Information Retrieval, Database, Human Computer Interaction, and Artificial Intelligence. Our task is to predict the category of a paper based on its title and abstract.

\textbf{WikiCS}~\cite{mernyei2020wiki} dataset is a Wikipedia-based dataset designed for benchmarking Graph Neural Networks. It is constructed from Wikipedia categories, specifically featuring 10 classes corresponding to branches of computer science, exhibiting very high connectivity. The node features are derived from the text of the corresponding articles. We obtain the raw texts of each node from https://github.com/pmernyei/wiki-cs-dataset.

\textbf{OGBN-ArXiv}~\cite{ogb} dataset is a directed graph representing the citation network among all computer science arXiv papers indexed by MAG~\cite{wang2020microsoft}. Each node corresponds to an arXiv paper, and directed edges indicate citations from one paper to another. The objective is to predict the 40 subject areas of arXiv CS papers, such as cs.AI, cs.LG, and cs.OS. These subject areas are manually determined and labeled by the paper's authors and arXiv moderators.

\textbf{Arxiv-2023}, proposed in TAPE~\cite{he2023explanations}, is a directed graph illustrating the citation network among all computer science arXiv papers published in 2023 or later. Like OGBN-ArXiv, each node represents an arXiv paper, and directed edges denote citations from one paper to another. The objective is to predict the 40 subject areas of arXiv CS papers, including cs.AI, cs.LG, and cs.OS. These subject areas are manually determined and labeled by the paper's authors and arXiv moderators.

\textbf{OGBN-Products}~\cite{ogb} is characterized by a substantial scale, comprising 2 million nodes and 61 million edges. We utilize a node sampling strategy, following TAPE~\cite{he2023explanations}, to obtain a subgraph containing 54k nodes and 74k edges, resulting in the OGBN-Products(subset) dataset. Each node in this dataset represents products sold on Amazon, and edges between two products indicate that the products are purchased together. The task involves predicting the category of a product in a multi-class classification setup, using the 47 top-level categories as target labels.

\textbf{Ele-Photo}~\cite{yan2023comprehensive} is derived from the Amazon-Electronics dataset~\cite{ni2019justifying}. The nodes in the dataset represent electronics-related products, and edges between two products indicate frequent co-purchases or co-views. The label for each dataset corresponds to the three-level label of the electronics products. User reviews on the item serve as its text attribute. In cases where items have multiple reviews, the review with the highest number of votes is utilized. For items lacking highly-voted reviews, a user review is randomly chosen as the text attribute. The task involves classifying electronics products into 12 categories.

\section{Baselines}\label{app:baselines}
In this section, we provide a detailed introduction to the baselines used:
\begin{itemize}
    \item Traditional GNNs: We adopt three simple but widely used GNN models in this work, \ie, GCN~\cite{gcn}, SAGE~\cite{sage}, GAT~\cite{gat}.
    \item GraphFormers~\cite{yang2021graphformers} is a graph transformer nested with GNN in each layer, originally designed for link prediction tasks. Here, $\text{GraphFormers}^\star$ refers to our modification of their official codes\footnote{https://github.com/microsoft/GraphFormers} for node classification tasks.
    \item NodeFormer~\cite{wu2022nodeformer} is an efficient graph transformer for large graphs which develops a kernelized Gumbel-Softmax~\cite{gumbel} operator.
    \item Fintuned LMs: We adopt three widely used pre-trained language models: BERT~\cite{kenton2019bert}, SentenceBERT~\cite{reimers2019sentence}, and DeBERTa~\cite{he2020deberta}. The parameters of these models are fully fine-tuned in our experiments.
    \item GIANT~\cite{giant} is a cascading structure method with two stages: pretraining LMs and training GNNs. In the first stage, it enhances node representations by integrating structural information into LM pre-training. Then fine-tuned LM-generated node features serve as initial features for GNN training. We use their official code\footnote{https://github.com/amzn/pecos/tree/mainline/examples/giant-xrt} for reproducing experiments.
    \item GLEM~\cite{glem} is an effective framework that fuses large language models and GNNs in the training phase through a variational EM framework. We used the source code\footnote{https://github.com/AndyJZhao/GLEM} provided in the original paper. Notably, both LMs and GNNs can be utilized for predictions in GLEM, so we report results for both.
    \item TAPE~\cite{he2023explanations} utilizes LLMs, like ChatGPT~\cite{openai2023gpt4}, to generate pseudo labels and explanations for textual nodes. Then it will finetune PLMs with the generated content and original texts. The enhanced features, derived from the fine-tuned PLMs, are used as initial node features for training GNNs. We use the provided features in their official code\footnote{https://github.com/XiaoxinHe/TAPE}.
    \item SimTeG~\cite{duan2023simteg} is also a cascading structure method tailored for textual graphs. It employs a two-stage training paradigm, initially fintuning language models and subsequently training GNNs. We use their official code\footnote{https://github.com/vermouthdky/SimTeG} for conducting experiments.
    \item PEFT methods: We incorporate four widely used PEFT methods, namely LoRA~\cite{lora}, IA3~\cite{IA3}, Prompt Tuning~\cite{prompt-tuning}, and LST~\cite{lst}. These methods involve fine-tuning large language models to showcase experimental results on textual graphs.
\end{itemize}

\section{Hyper-parameters}\label{app:hyper}
The hyperparameters for PEFT methods are outlined in Table~\ref{tab:app_peft}, determined through grid search for optimal results. Specifically, training epochs vary within $\{4, 6, 8, 10\}$ for small datasets (\eg, cora, citeseer, wikics), $\{1, 2, 4\}$ for medium datasets (\eg, arxiv-2023, products, and photo), and 1 for the large dataset (ogbn-arxiv). Learning rate is explored in $\{0.01, 0.001, 0.0001, 0.00001\}$. 
The model configuration for the GNN used in G-Ladder is detailed in Table~\ref{tab:app_baseline}.

\section{Applying \ours on Other LMs}\label{app:e5}
Our method is versatile and applicable to any language models. In this section, we demonstrate the effectiveness of our approach by applying it to another renowned language model, namely e5-large~\cite{e5-large}. 

Table~\ref{tab:e5} reveals that \ours outperforms other PEFT methods, and \oures achieves comparable performance with \ours while significantly reducing inference time. Similar observations are found in Section~\ref{sec:performance}.

\begin{table}[htp]

    \centering
    \scalebox{0.9}{
    \begin{tabular}{l|ccccccc}
    \toprule
                  & Cora         & CiteSeer   & WikiCS       \\ \midrule
e5-large (\faLock)    & ---     & ---     & ---       \\
\quad+LoRA          & 79.89±1.95 & 73.13±1.84 & 78.70±0.48   \\
\quad+IA3           & 77.60±1.58 & 71.94±1.47 & 74.68±0.40   \\
\quad+Prompt Tuning & 72.92±1.76 & 70.31±1.93 & 64.57±2.88 \\
\quad+LST           & 73.76±1.30 & 71.16±1.57 & 64.78±1.21  \\
\rowcolor{Gray}
\quad+\our          & 90.59±0.44 & 77.87±1.21 & 80.15±0.57\\
\rowcolor{Gray}
\quad+\oure    & 90.18±0.46 & 77.41±1.02 & 79.50±0.31 \\
\bottomrule
    \end{tabular}}
    \caption{Experimental results of PEFT methods and \ours on e5-large model. \faLock\ denotes freezing the parameters of LMs.}
    \label{tab:e5}
\end{table}

\section{Different Message Passing in G-Ladders}\label{app:message}
Various message passing mechanisms can be employed in our G-Ladders. For simplicity, we utilize GCN, SAGE, and GAT in this context. More advanced methods, such as RevGAT, can be considered for future exploration. We tune the hyper-parameters based on SAGE and apply the same hyper-parameters on other GNNs, the used hyper-parameters can be found in Table~\ref{tab:gladder}.

Table~\ref{tab:gladder} presents the experimental results, demonstrating comparable outcomes across different message passing mechanisms.

\begin{table}[htp]
    \centering
    \begin{tabular}{c|ccc}
    \toprule
       Methods & Cora & CiteSeer & WikiCS \\ \midrule
        MLP    & 83.92±1.39 & 91.00±0.95&  92.31±0.07 \\
        GCN    & 90.22±0.89 & 92.93±1.36&  93.63±0.24      \\
        SAGE   & 88.25±0.88 & 91.68±1.08    &   95.93±0.20     \\
        GAT    & 89.70±1.72 & 91.95±0.90 & 93.25±0.13       \\ \midrule
        \our   & \textbf{94.23±0.77} & \textbf{95.65±0.69}& \textbf{98.30±0.17}  \\ 
        
        \bottomrule
    \end{tabular}
    \caption{Link prediction performance, as evaluated by AUC metric.}
    \label{tab:link}
\end{table}

\section{Link Prediction}\label{app:link}
Our method is not limited to node classification tasks, it can also be applied to edge-level or graph-level tasks. In this section, we conduct experiments on link prediction tasks. We split existing edges into train:val:test=0.85:0.05:0.1 for all datasets. The reported results include the mean AUC with standard deviation across 5 different random seeds. For simplicity, we use traditional GNN methods as baselines and present the results of our method.

According to Table~\ref{tab:link}, \our outperforms traditional GNN models by a considerable margin on the three datasets. In particular, \ours exhibits more than a 4\% absolute improvement over GCN on the Cora dataset. These statistics underscore the effectiveness of our method in edge-level tasks.

\begin{table*}[ht]

    \centering
    \begin{tabular}{c|ccccccc}
    \toprule
                  & Cora     & CiteSeer& WikiCS   & OGBN-ArXiv & ArXiv-2023 & OGBN-Products & Ele-Photo \\ \midrule
    \# Epochs     & 4        & 8       & 10      &     1        &        2 &   1        &          2\\
    Learning rate & 1e-3     & 1e-2    & 1e-2    &     1e-3     &        1e-2&   1e-3     &          1e-3\\
    Batch size    & 16       & 16    & 16      &     16       &        16&   16       &          16\\
    Optimizer     & AdamW    & AdamW   & AdamW   &     AdamW    &        AdamW&        AdamW&          AdamW\\
Gradient acc. steps & 2      & 2    & 2       &     2        &        2&   2        &          2\\ 
    \bottomrule
    \end{tabular}
    \caption{Hyper-parameters for PEFT of LLM baselines.}
    \label{tab:app_peft}
\end{table*}

\begin{table*}[ht]

    \centering
    \begin{tabular}{c|ccccccc}
    \toprule
                  & Cora     & CiteSeer& WikiCS   & OGBN-ArXiv & ArXiv-2023 & OGBN-Products & Ele-Photo \\ \midrule

    \# Hidden size& 64       & 64      & 64       &     64    &        64  &        64  &          64\\
    \# Layers     & 2        & 1       & 1        &     1      &        1   &        1   &          2\\
    Norm          & ID       & ID      & ID       &     LN     &        LN  &        LN  &          ID   \\
    Activation    & ELU      & ELU     & ReLU     &     ELU    &        ELU &        ELU  &          ELU  \\ 
    Dropout       & 0.5      & 0.5     & 0.5      &     0.2    &         0.2&        0.2 &          0.5  \\ 
    \# Epochs     & 200      & 200     & 200      &     200    &        200 &        200&           200\\
    Learning rate & 5e-5     & 5e-5    & 1e-3     &     1e-4   &        1e-4&        1e-2&          1e-3\\
    Optimizer     & AdamW    & AdamW   & AdamW    &     AdamW  &       AdamW&        AdamW&          AdamW\\
    Weight decay  & 5e-4     & 5e-4    & 1e-4     &     5e-3   &        5e-4&        5e-4&          5e-4\\ 
    Early stop    & True     & True    & True     &     True   &        True&        True&          True \\ 
    Patience      & 20       & 20      & 20       &     5      &          5&          20 &          20  \\ 
    Sampler       & RWR      & KHop    & KHop     &     KHop   &        KHop&        KHop &          KHop  \\ 
    \bottomrule
    \end{tabular}
    \caption{Hyper-parameters for GNN baselines. `ID' means no norm layer(Identity), `LN' denotes Layer Normalization. For sampler, `RWR' means random walk sampler with restart, and `KHop' is one-hop ego graph sampler.}
    \label{tab:app_baseline}
\end{table*}

\begin{table*}[ht]
    \centering
    \begin{tabular}{cc|cccccc}
    \toprule
                &         & Cora       & CiteSeer    & WikiCS     & OGBN-ArXiv & ArXiv-2023 & OGBN-Products \\  \midrule
\multirow{3}{*}{\shortstack{LLaMA2-7B\\+\our}}   
                & GCN     & 90.63±0.38 & 77.40±0.65 & 80.91±0.74 & 75.77±0.38 & 77.66±0.27 & 78.94±0.83    \\
                & SAGE    & 91.48±0.32 & 78.46±0.49 & 81.56±0.97 & 76.02±0.29 & 79.12±0.16 & 80.05±0.45    \\
                & GAT     & 90.33±0.38 & 77.81±0.72 & 81.43±0.83 & 75.58±0.44 & 78.27±0.12 & 78.42±1.27    \\ 
\bottomrule
    \end{tabular}
    \caption{Experimental results of different GNNs used in G-Ladders.}
    \label{tab:gladder}
\end{table*}

\end{document}